%% file: main.tex
\documentclass[letterpaper, 10 pt, journal, twoside]{ieeetran}

\usepackage{amsmath,amsfonts}
\usepackage{array}
\usepackage[caption=false,font=normalsize,labelfont=sf,textfont=sf]{subfig}
\usepackage{textcomp}
\usepackage{stfloats}
\usepackage{url}
\usepackage{verbatim}
\usepackage{graphicx}
\usepackage{cite}
\usepackage{amssymb}

\usepackage{color, colortbl}
\usepackage[colorlinks,bookmarksnumbered,citecolor=orange,urlcolor=orange]{hyperref}
\graphicspath{{Figures}}
\DeclareGraphicsExtensions{.pdf,.png}
\usepackage{bigstrut}
\usepackage[english]{babel}

\usepackage{csquotes}

\usepackage[T1]{fontenc}
\usepackage{algorithm, setspace}
\usepackage{algpseudocode}
\usepackage{multirow}
\usepackage{float}
\usepackage{xcolor}
\usepackage{hyperref}
 \hypersetup{
     colorlinks=true,
     linkcolor=orange,
     filecolor=orange,
     citecolor=orange,      
     urlcolor=orange,
     }

\usepackage{dsfont}

\newcommand{\p}[1]{\smallskip \noindent \textbf{{#1}.}}
\newcommand{\eq}[1]{Equation~(\ref{eq:#1})}
\newcommand{\fig}[1]{Figure~\ref{fig:#1}}
\newcommand{\figs}[2]{Figures~\ref{fig:#1} and \ref{fig:#2}}

\usepackage{balance}

\title{Encouraging Human Interaction with Robot Teams: \\ Legible and Fair Subtask Allocations
}

\author{Soheil Habibian,~\IEEEmembership{Student Member,~IEEE,} and Dylan P. Losey
\thanks{Soheil Habibian and Dylan P. Losey are members of the Collaborative Robotics Lab (\href{https://collab.me.vt.edu/}{Collab}), Dept. of Mechanical Engineering, Virginia Tech, Blacksburg, VA 24060 USA. \texttt{\{habibian, losey\}@vt.edu}}%
}

\begin{document}
\maketitle

\begin{abstract}

Recent works explore collaboration between humans and teams of robots. These approaches make sense if the human is already working with the robot team; but how should robots encourage nearby humans to join their teams in the first place? Inspired by economics, we recognize that humans care about more than just team efficiency --- humans also have biases and expectations for team dynamics. Our hypothesis is that the way inclusive robots divide the task (i.e., how the robots split a larger task into subtask allocations) should be both \textit{legible} and \textit{fair} to the human partner. In this paper we introduce a bilevel optimization approach that enables robot teams to identify high-level subtask allocations and low-level trajectories that optimize for legibility, fairness, or a combination of both objectives. We then test our resulting algorithm across studies where humans \textit{watch} or \textit{play} with robot teams. We find that our approach to generating legible teams makes the human’s role clear, and that humans typically prefer to join and collaborate with legible teams instead of teams that only optimize for efficiency. Incorporating fairness alongside legibility further encourages participation: when humans play with robots, we find that they prefer (potentially inefficient) teams where the subtasks or effort are evenly divided. See videos of our studies here: \url{https://youtu.be/cfN7O5na3mg} 

\end{abstract}







\input{intro}
\input{related}
\input{problem}
\input{method}
\input{mturk}

\input{user-study}
\input{conclusions}


\balance
\bibliographystyle{IEEEtran}
\bibliography{bibliography}

\end{document}

%% file: intro.tex
\section{Introduction}

\begin{figure}[t]
\centerline{\includegraphics[width=.9\columnwidth]{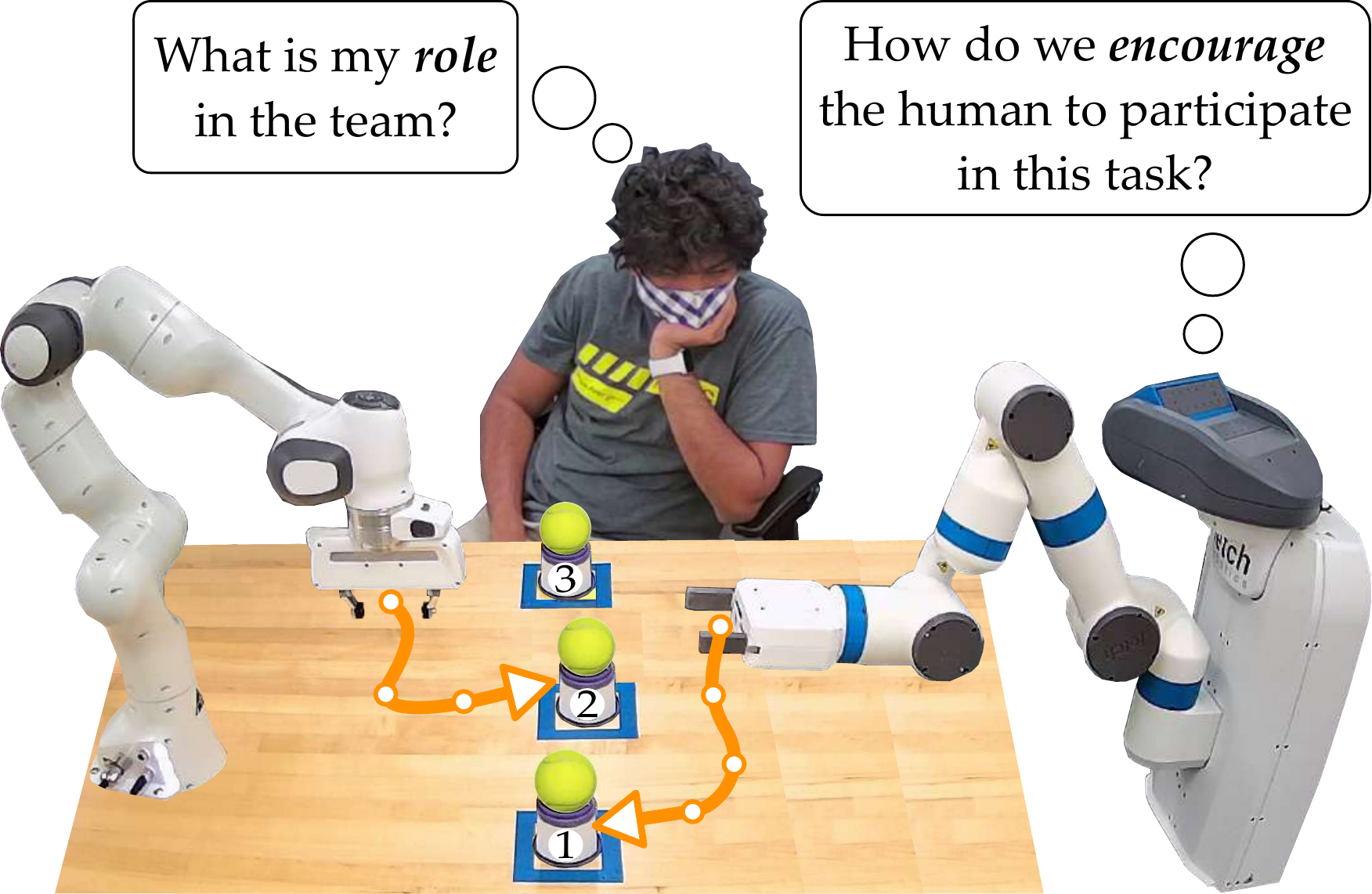}}
\caption{Robot team clearing a cluttered table. A human is thinking about joining this team, but the human does not know \textit{a priori} which ball they should grab. We find that robot teams can encourage humans to collaborate by purposefully dividing the task into \textit{legible} and \textit{fair} subtask allocations.}
\label{fig:front}
\vspace{-1.5em}
\end{figure}

\IEEEPARstart{I}{magine} sitting next to the team of robot arms in \fig{front}. These robots are working together to clear a cluttered table: there are multiple tennis balls that the team needs to remove, and each robot arm is reaching to grab a different ball. You know the robots' high-level task, but you do not know how the team will complete that task --- or what you can do to help. How should the robot team \textit{divide} and perform the task to \textit{encourage} you to join in and collaborate?

Recent work on multi-agent reinforcement learning \cite{gupta2017cooperative, foerster2019bayesian, lowe2017multi} and decentralized control \cite{culbertson2021decentralized} explores teams composed entirely of autonomous agents. Other current research brings a human into these teams: this includes learning to play with humans \cite{carroll2019utility}, influencing teams of humans \cite{kwon2019influencing}, inferring the teammates' roles \cite{wu2021too}, and forming new teams without pre-coordination \cite{stone2010ad}. Viewed together, these prior works take a step towards enabling teams of robots to collaborate with a human that is \textit{already} participating in the team \cite{sebo2020robots}.

But how should robots encourage humans to join their teams in the first place? We hypothesize that the way the robots divide the task among agents --- and the way each robot performs its subtask --- will affect the human's willingness to collaborate with the robot team. Prior work in behavioral economics \cite{clark1998comparison, fehr1999theory, eshghi2017mathematical} suggests that task efficiency is not the only factor that encourages humans to form teams. Instead of treating human teammates like robot partners, robots must account for human biases and expectations:
\begin{center}\vspace{-0.4em}
\textit{Humans are inclined to participate in teams \\ where their roles are \emph{legible} and \emph{fair}.}\vspace{-0.4em}
\end{center}
Returning to our motivating example from \fig{front}, there are many equally efficient ways for the robots to grab the balls and clear the table. But robots which apply our insight reach for balls $1$ and $2$: by moving for the other end of the table they make the human's role clear, and evenly divide the subtasks among the three teammates. Accordingly, in this paper we study two axes for subtask allocation: \textit{legibility} and \textit{fairness}. We experimentally test how both of these factors encourage participation when humans are watching robot teams and when humans join in and play with those teams. Our work is motivated by mixed-autonomy settings (such as factory floors) where we want to facilitate human-robot collaboration.

Overall, we make the following contributions:

\p{Formalizing Legible and Fair Subtask Allocations} We first introduce a bilevel optimization approach that enables centralized robot teams to identify high-level subtask allocations and low-level trajectories. We then formulate legible and fair allocations when the human is watching and playing, and incorporate both into our bilevel optimization.

\p{Encouraging Humans to Join Robot Teams} We conduct two online user studies where humans watch robot teams. In the first user study we find that humans prefer legible teams over teams that only optimize for task performance. In the second user study we also incorporate fairness, and compare legible teams to teams that are both legible \textit{and} fair.

\p{Encouraging Humans to Keep Collaborating} We next evaluate our approach in an in-person user study where humans collaborate with two robot arms. We find that humans prefer to keep working with teams of robots that optimize for legibility instead of efficiency. We also find that fairness has a statistically significant impact: legible and fair teams better encourage collaboration than teams which are only legible.

%% file: related.tex
\section{Related Work}

\noindent \textbf{Multi-Agent Teams.} Recent works enable teams composed purely of autonomous agents to perform collaborative tasks \cite{gupta2017cooperative, foerster2019bayesian, lowe2017multi, culbertson2021decentralized}. However, it is still challenging for these autonomous teams to incorporate humans \cite{stone2010ad, carroll2019utility, parekh2022rili}. One common method for facilitating human-robot teams is introducing subtasks (or roles), and then assigning these subtasks to the robots and humans \cite{kwon2019influencing, johannsmeier2016hierarchical, rahman2018mutual}. Gombolay \textit{et al.} \cite{gombolay2015decision} find that humans prefer to work in teams where robots assign subtasks, and Wang \textit{et al.} \cite{wu2021too} suggest that humans can infer the subtasks of autonomous agents by observing their motion. We build on these approaches by similarly using subtasks. But unlike prior works --- where often there is only one robot, and the human and robot have already formed a team \cite{sebo2020robots} --- here we focus on \textit{bringing} one human into a team with \textit{multiple} robots.

\p{Legible Interaction} Robots can leverage their behavior to implicitly communicate goals, objectives, or uncertainty to human partners \cite{liu2018goal, habibian2022here, hellstrom2018understandable}. Most relevant is research by Roncone \textit{et al.} \cite{roncone2017transparent}, which indicates that transparency is key when humans and robots are deciding subtask allocations. But while prior work explores how one robot should convey its intent to the human \cite{dragan2013legibility}, how should \textit{teams} of robots communicate their overall allocation? We will extend legibility from dyads to teams, and optimize the team's behavior so that human onlookers can infer their intended subtask(s).

\p{Fairness in Human-Robot Teams} Research from psychology and economics indicates that humans have expectations for the teams they join: in particular, humans expect those teams to be fair \cite{barrick1998relating, rabin1993incorporating}. While prior work largely focuses on human-human teams, state-of-the-art studies extend those same principles to human-robot teams \cite{claure2020multi, chang2020defining, chang2021unfair}. For instance, Claure \textit{et al.} \cite{claure2020multi} impose a fairness constraint on resource distribution in multi-armed bandits, and find that this fairness impacts the the human's trust in the system. Other works explore human perceptions of fairness in terms of assigned workload, member capability, and task type \cite{chang2020defining, chang2021unfair}. In this paper we leverage definitions of fairness that are consistent with prior works, but we now focus on how the fairness of subtask allocations affects the human's willingness to \textit{join} robot teams.

%% file: problem.tex
\section{Problem Setting} \label{sec:problem}

We explore scenarios where a robot team is collaborating to perform a task, and these robots want to encourage nearby humans to join in and participate. As our \textit{running example}, consider the two robot arms that are clearing a cluttered table in \fig{front}. We introduce additional structure by dividing the overall task into \textit{subtasks}. These subtasks could be steps towards a larger goal --- e.g., removing one tennis ball from the cluttered table --- or they could be roles within the task --- e.g., leader and follower. We assume that the team of robots are centralized: they communicate with one another in real time and share a common controller.

\p{MDP with Subtasks} From the robots' perspective this is an instance of a Markov decision process (MDP) with subtasks: $\langle \mathcal{S}, \mathcal{A}, T, R, \gamma, \mathcal{T} \rangle$. Let $s \in \mathcal{S}$ denote the system state, let $a \in \mathcal{A}$ denote the system action, and let $s^{t+1} = T(s^t, a^t)$ be the system dynamics. We emphasize that the state $s$ and action $a$ contain the combined states and actions of every robot teammate: returning to our running example, the action $a$ is the joint velocities of both robot arms. At each timestep $t$ the robots take actions to interact with the environment. We write the team's sequence of states and actions up to time $t_{K}$ as a \textit{trajectory} $\xi = \{(s^0, a^0), \ldots, (s^{t_{K}}, a^{t_{K}})\}$.

The team of robots is collaborating to perform a task. We capture this objective through the sparse reward function $R(s) = \mathds{1} \{ \text{task solved in } s\}$, which indicates whether or not the task is complete at state $s$ with discount factor $\gamma \in [0, 1)$. But we also break the overall task into subtasks: let $\tau \in \mathcal{T}$ be a subtask, and let $\mathcal{T}$ be the finite set of required subtasks. By completing all of these subtasks the team reaches the goal state and receives reward $R = 1$. In our running example the task is to clear the table, and the three subtasks $\tau_1$, $\tau_2$, and $\tau_3$ are removing the tennis balls marked $1$, $2$, and $3$.

\p{Allocations} Introducing subtasks brings with it a challenge: how should the team of robots divide these subtasks among themselves and the nearby human? Let $\theta \in \Theta$ denote an \textit{allocation}. This allocation $\theta$ determines which subtask(s) each robot will perform and which subtask(s) the robots want the human to complete. Our running example has three subtasks $\mathcal{T} = \{\tau_1, \tau_2, \tau_3\}$. Here one allocation could be ${\theta = \{(human: \tau_2), (robot_1: \tau_3), (robot_2: \tau_1)\}}$, indicating that the human is assigned subtask $\tau_2$ and the robots will perform subtasks $\tau_1$ and $\tau_3$. It is also possible for an agent to be assigned either \textit{no} subtasks or \textit{multiple} subtasks: for instance, ${\theta = \{(human: \emptyset), (robot_1: \tau_1, \tau_2), (robot_2: \tau_3)\}}$. Moving forward we will use $\mathcal{T}_{i}(\theta)$ to refer to set of subtask(s) assigned to the $i$-th agent under allocation $\theta$.

\p{Fixed Robots} The robots assume that the human will follow their chosen allocation $\theta \in \Theta$. Put another way, the robots are \textit{fixed}: they select and execute a single allocation during each interaction, and do not switch their allocation in response to the human \cite{liu2018goal}. We make this assumption in order to isolate how the human responds to the robots, and avoid entangling this with how the robots respond to the human.

%% file: method.tex
\section{Optimizing for Legible and Fair Allocations} \label{sec:method}

Given the formulation from Section~\ref{sec:problem}, we search for a subtask allocation $\theta$ that encourages nearby humans to join in and participate with the robot team. We hypothesize that robots encourage human participation by optimizing along two axes: \textit{legibility} and \textit{fairness}. Here legibility refers to how clearly the robots convey their allocation $\theta$ to the human: e.g., which tennis balls are the robots clearing, and which ball should the human reach for? Fairness captures how the evenly subtasks are divided among the teammates: e.g., is the human expected to remove the same number of balls as each robot teammate? Below we formally define legibility and fairness, and introduce a bilevel optimization approach for identifying legible and fair allocations and trajectories.

\p{Bayesian Inference} At the start of the interaction the human is not sure what subtasks each agent should complete; however, the human can infer $\theta \in \Theta$ based on the robots' behavior $\xi$. Applying Bayes' theorem:
\begin{equation} \label{eq:B1}
    P(\theta \mid \xi) = \frac{P(\xi \mid \theta) \cdot P(\theta)}{\sum_{\theta' \in \Theta}P(\xi \mid \theta') \cdot P(\theta')}
\end{equation}
where $P(\theta)$ is the prior over allocations and $P(\xi \mid \theta)$ is the likelihood of allocation $\theta$ given team trajectory $\xi$. Since the trajectory $\xi$ is composed of conditionally independent state-action pairs $(s, a)$, we rewrite the likelihood function:
\begin{equation} \label{eq:B2}
    P(\xi \mid \theta) = \prod_{(s, a) \in \xi} P\big((s, a) \mid \theta\big)
\end{equation}
We assume that the human views the team as a Boltzmann-rational agent. This model --- commonly used in robotics \cite{ziebart2008maximum, jeon2020reward} and economics \cite{luce2012individual} --- assigns higher likelihood to actions that lead to increased long-term reward:
\begin{equation} \label{eq:B3}
    P(\xi \mid \theta) \propto \exp{\sum_{(s, a) \in \xi}Q_{\theta}(s, a)}
\end{equation}
Here $Q_{\theta}(s,a)$ is the cumulative reward of taking action $a$ in state $s$ and optimally completing allocation $\theta$ thereafter. Within our running example the allocation $\theta$ tells each agent which ball to reach for, and $Q_{\theta}(s,a) = -d(T(s, a), g_{\theta})$ is the negative distance between the next state and the goal state $g_{\theta}$. Combining Equations (\ref{eq:B1})--(\ref{eq:B3}) provides a model that the robots can evaluate for $P(\theta \mid \xi)$.

\subsection{Efficient Allocations}

Before considering legibility or fairness, we start with an \textit{efficient} baseline. This team of robots optimizes purely for task performance: put another way, the robots select an allocation $\theta \in \Theta$ that maximizes their long-term reward and completes the task as quickly as possible. In our experiments we select allocations that are noisily-optimal: $P(\theta_{\mathcal{E}\text{ff}} \mid s) \propto V_{\theta}(s)$, where $V_{\theta}(s)$ is cumulative reward for starting at $s$ and completing the task using allocation $\theta$. Within our running example this produces allocations $\theta_{\mathcal{E}\text{ff}}$ where each agent removes one tennis ball, and the robots reach directly towards their allocated ball.

\subsection{Legible Allocations}

We next leverage $P(\theta \mid \xi)$ to optimize for legible --- but potentially inefficient --- allocations. We emphasize that \textit{legible allocations} are different from \textit{legible motions} \cite{dragan2013legibility}. Within legible motions the team of robots is selecting a trajectory $\xi$ that communicates one specific allocation $\theta$, i.e., the robots are maximizing $P(\theta \mid \xi)$. Legible allocations require another level of optimization: now the team of robots must not only find the best way to convey a given $\theta$, but they must also determine which allocation $\theta \in \Theta$ they are able to convey most clearly. This results in a bilevel optimization problem across continuous trajectories and discrete allocations.

\p{Watching} When the human is watching a team of robots --- and is not participating in the task themselves --- we identify legible allocations and trajectories by optimizing:
\begin{equation} \label{eq:L1}
    \big(\theta_{\mathcal{L}eg}, ~\xi_{\mathcal{L}eg}\big) = \underset{\theta \in \Theta, \xi \in \Xi}{\text{argmax}} ~ P(\theta \mid \xi)
\end{equation}
Here the lower-level optimization problem is finding the trajectory $\xi$ that maximizes the likelihood of $\theta$, and the upper-level optimization problem iterates through each choice of $\theta \in \Theta$ to find the allocation that maximizes $P(\theta \mid \xi)$. 


\p{Playing} Legible allocations change when the human is joining in to collaborate with the robots. Now the human is not concerned with the overall allocation for every teammate --- instead, the human only needs to know their own subtask(s). Returning to our running example, it does not matter which robot is picking up ball $1$ or ball $3$; the human just needs to know that their job is picking up ball $2$. Hence, we introduce $\mathcal{H}(\theta)$, the set of all allocations $\theta' \in \Theta$ where the human has the same subtasks. More formally, $\mathcal{H}(\theta) = \{\theta' \mid \mathcal{T}_{\mathcal{H}}(\theta') = \mathcal{T}_{\mathcal{H}}(\theta)\}$. We then sum across this set:
\begin{equation} \label{eq:L2}
    \big(\theta_{\mathcal{L}eg}, ~\xi_{\mathcal{L}eg}\big) = \underset{\theta \in \Theta, \xi \in \Xi}{\text{argmax}} ~ \sum_{\theta' \in \mathcal{H}(\theta)} P(\theta' \mid \xi)
\end{equation}
Intuitively, \eq{L2} marginalizes out the robots' specific roles, leading to an allocation $\theta_{\mathcal{L}eg}$ where the human can best infer what subtask(s) they are meant to do.

\begin{figure*}[t]
	\begin{center}
		\includegraphics[width=2\columnwidth]{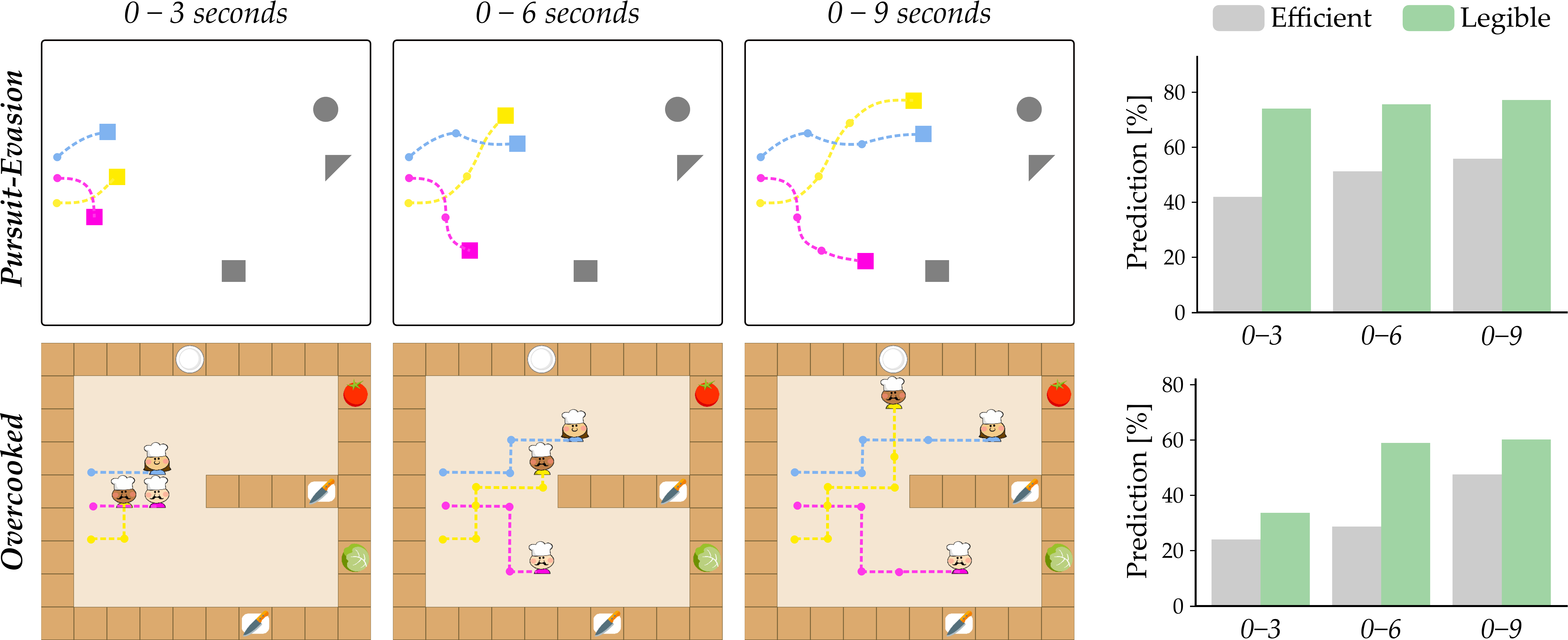}
		\caption{Environments and results from our online user study. (Left) Participants watched videos of simulated robot teams in \textit{Pursuit-Evasion} and \textit{Overcooked} environments. We showed videos of the team's behavior during the first three seconds, six seconds, and nine seconds: here dotted lines depict an example of the trajectories the agents traveled in each video snippet. (Right) Based on these videos participants predicted the team's subtask allocation. Users more accurately predicted the subtasks of teams that optimized for \textbf{Legible} allocations as compared to \textbf{Efficient} teams that optimized for task performance.} 
		\label{fig:amt_pred}
	\end{center}
	\vspace{-1.5em}
\end{figure*}

\subsection{Fair Allocations}

Optimizing for legibility enables the human to understand their role within the team. But just because their subtask(s) are clear does not mean that the human will want to complete these subtask(s). We therefore introduce \textit{fairness} as a second axis for encouraging collaboration. Our approach is agnostic to the specific function used to quantify fairness, but in our experiments we tested \textit{two definitions} from related works on economics \cite{clark1998comparison, fehr1999theory} and robotics \cite{claure2020multi, chang2020defining}. Let $f_i(\theta, \xi)$ be the fairness for agent $i$ given allocation $\theta$ and trajectory $\xi$. Our first approach to fairness is \textit{equality of allocation}:
\begin{equation} \label{eq:F1}
    f_i(\theta, \xi) = -\bigg|~\frac{|\mathcal{T}|}{N} - \big|\mathcal{T}_i(\theta)\big|~\bigg|
\end{equation}
Here $N$ is the total number of teammates, and allocation $\theta$ is fair if it assigns an equal number of subtasks to each agent. Our second approach is \textit{equality of effort}:
\begin{equation} \label{eq:F2}
    f_i(\theta, \xi) = -\bigg|~\frac{d(\xi)}{N} - d_i(\xi)~\bigg|
\end{equation}
where $d(\xi)$ is the overall distance the team must travel under trajectory $\xi$, $d_i(\xi)$ is the distance the $i$-th agent travels, and an allocation is fair if each team member travels the same distance. Regardless of our chosen definition for $f$, the team of robots again solves a bilevel optimization problem to identify allocations that are now \textit{both fair} and \textit{legible}.

\p{Watching} When the human is watching a team they are not compelled to take the perspective of any specific agent. Put another way, the allocation should be fair for \textit{every} agent. We therefore solve for fair and legible allocations by optimizing:
\begin{equation} \label{eq:F3}
    \big(\theta_{\mathcal{F}air}, ~\xi_{\mathcal{F}air}\big) = \underset{\theta \in \Theta, \xi \in \Xi}{\text{argmax}} ~ \sum_{i = 1}^{N}f_i(\theta, \xi) + P(\theta \mid \xi)
\end{equation}
We note that this builds on \eq{L1}, and now encourages the robots to leverage fair allocations which the human can correctly interpret.

\p{Playing} The human's perspective changes when they collaborate and actively participate with the team of robots. Here we hypothesize that the human focuses on the how fair the allocation is for \textit{themselves} --- e.g., is the human being asked to pick up and move more tennis balls than both of the robots? Consistent with \eq{L2}, we optimize:
\begin{equation} \label{eq:F4}
    \big(\theta_{\mathcal{F}air}, ~\xi_{\mathcal{F}air}\big) = \underset{\theta \in \Theta, \xi \in \Xi}{\text{argmax}} ~ f_{\mathcal{H}}(\theta, \xi) + \sum_{\theta' \in \mathcal{H}(\theta)} P(\theta' \mid \xi)
\end{equation}
where $f_{\mathcal{H}}$ is the fairness for the human teammate.


%% file: mturk.tex
\section{Encouraging Participation when Watching}\label{sec:mturk}

Before describing this experiment we first want to outline our user studies in Section~\ref{sec:mturk} and \ref{sec: user_study}. In each section we will compare robot teams that optimize for efficiency, robot teams that optimize for legibility, and robot teams that optimize for legibility and fairness. The key difference between these two sections is whether the human is \textit{watching} (Section~\ref{sec:mturk}) or \textit{playing} (Section~\ref{sec: user_study}) with the robot team. Both aspects are important: we want to determine which types of allocations draw the human into the robot team (\textit{watching}) and encourage the human to continue collaborating with that team (\textit{playing}). Moreover, recall that we have different equations for legibility and fairness in each context: when the human is watching they can consider the task from the perspective of any agent, but once the human joins they must focus on the legibility and fairness of their own subtask.

Here we start with \textit{watching}. We perform two separate online user studies where humans observe multi-agent teams in two simulated environments (see \fig{amt_pred}). In the first study we compare robot teams that only consider efficiency to robot teams that optimize for legibility. We test whether these legible allocations actually make it easier for humans to predict the subtasks, and whether legible allocations encourage onlookers to join robot teams. In the second study we compare legible teams to teams that are both legible \textit{and} fair.

\p{Environments} We used two simulated environments from prior work on multi-agent teams and legible motion. Both environments had an overall task that was divided into subtasks $\mathcal{T}$ for the agents to complete. In \textit{Pursuit-Evasion} \cite{kwon2019influencing, dragan2013legibility} the state-action space is continuous and the multicolored agents are trying to reach the gray targets. Each agent has at least one subtask (i.e., $\mathcal{T}_i(\theta)\neq\emptyset$), and it is possible for multiple agents to share the same target. In \textit{Overcooked} \cite{wu2021too, carroll2019utility} agents operate in a discrete state-action space and their targets include ingredients (lettuce and tomato) or kitchen utensils (cutting board and plate). Here agents never shared the same target and an agent could be assigned no subtasks (i.e., $\mathcal{T}_i(\theta)=\emptyset$).

\p{Participants} We recruited $100$ total participants on Amazon Mechanical Turk (MTurk). Our experiment was only available to English speaking participants who had completed at least $100$ Human Intelligence Tasks (HITs) with a $99\%$ approval rating. Users had to correctly answer qualifying questions to ensure that they had read and understood our instructions before they could participate. We then divided these participants into two groups: $50$ users completed the study in Section~\ref{sec:watch_legibility} that compares efficiency and legibility, and the other $50$ users completed the study in Section~\ref{sec:watch_fairness} that compares legibility without fairness to legibility with fairness.

\subsection{Legibility} \label{sec:watch_legibility}

\begin{figure}[t]
\centerline{\includegraphics[width=1\columnwidth]{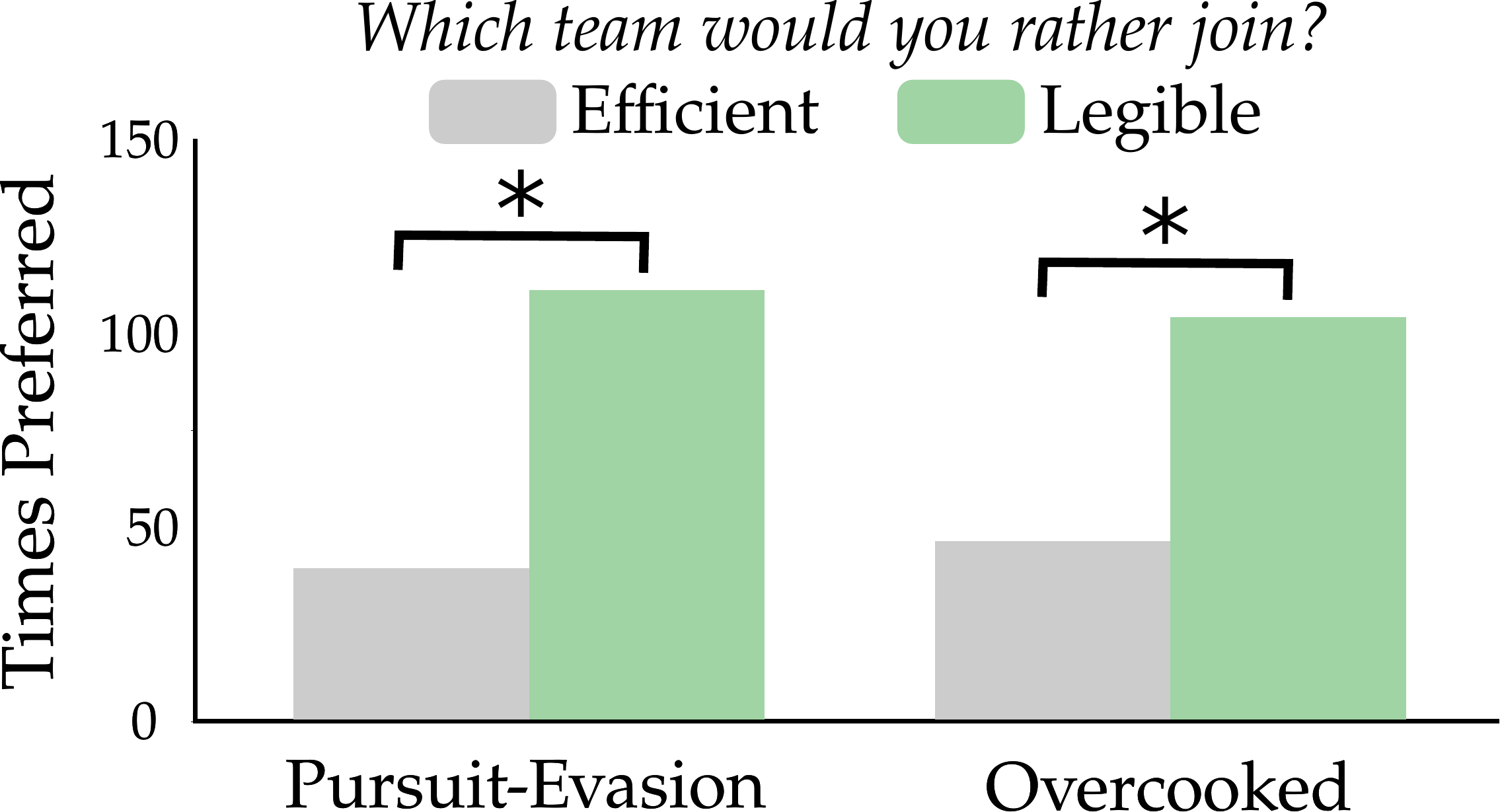}}
\caption{User preferences when watching \textbf{Efficient} and \textbf{Legible} robot teams. Fifty participants responded to three forced-choice comparisons by indicating which team they would prefer to work with in the future. For example, in \textit{Pursuit-Evasion}
users preferred \textbf{Legible} teams $111$ times and \textbf{Efficient} teams $39$ times. {Here $*$ denotes statistical significance ($p<.001$).}}


\label{fig:amt_pref_leg}
\vspace{-1.5em}
\end{figure}

Our baseline is a team of robots that optimize for efficiency: these robots choose allocations to complete the task as quickly as possible. Here we compare that baseline to a \textit{legible} robot team. Participants watch teams of agents complete tasks in our simulated environments. We test whether legible teams better convey their allocation to the human, and whether humans prefer to join these teams.

\p{Independent Variables} We compared two types of subtask allocations: \textbf{Efficient} and \textbf{Legible}. In \textbf{Efficient} the robots selected noisily-rational allocations $\theta_{\mathcal{E}\text{ff}}$. Under \textbf{Legible} the robots selected the allocation $\theta_{Leg}$ that optimizes \eq{L1} and attempts to reveal every agent's subtask. To better isolate legibility, we tested \textbf{Efficient} teams that \textit{were not} legible.

\p{Procedure} Our $50$ participants first watched videos of \textbf{Efficient} and \textbf{Legible} allocations (see \fig{amt_pred}). These videos showed the team's behavior after three, six, and nine seconds had passed. While watching these videos participants indicated which subtask they thought each agent was completing; e.g., based on the motion from $0-9$ seconds, a user might guess that the blue chef is reaching for the red tomato. Participants had to indicate their prediction at the current timestep before watching the next increment. Finally, participants were shown three side-by-side comparisons of \textbf{Efficient} and \textbf{Legible} teams. After watching both robot teams complete the task participants were asked to chose one team to join. We randomized the order of \textbf{Efficient} and \textbf{Legible} teams; participants were never told which allocations they were watching. 

\p{Dependent Measures} We counted the total number of times the participants correctly guessed the subtasks of all three agents when watching an \textbf{Efficient} allocation and when watching a \textbf{Legible} allocation. We also recorded the total number of times participants preferred \textbf{Efficient} allocations, and the total number of times participants preferred \textbf{Legible} allocations.

\p{Hypotheses} We had two hypotheses in this user study:

\p{H1} \textit{Human observers will more accurately predict the subtasks of teams that optimize for legible allocations.}

\p{H2} \textit{Humans will prefer to join teams that optimize for legible subtask allocations.}

\p{Results} The results of our first \textit{watching} survey are shown in \figs{amt_pred}{amt_pref_leg}. Across both \textit{Pursuit-Evasion} and \textit{Overcooked} environments, more participants correctly predicted the roles of each team member when watching \textbf{Legible} teams (\fig{amt_pred}). This prediction accuracy increases as the interaction unfolds: when the agents get closer to completing the task their subtasks become increasingly clear to the human. Overall, the participants' responses support \textbf{H1}.

Importantly, the \textit{legibility} of a team's allocation affected people's willingness to join that team. \fig{amt_pref_leg} displays the participant's preference when asked: \textit{``if you had to join team A or team B, which would you join?''} In both environments participants chose \textbf{Legible} teams more frequently. {Two Wilcoxon signed-rank tests showed that these differences were statistically significant in \textit{Pursuit-Evasion} ($Z=-5.879$, $p<.001$) or in \textit{Overcooked} ($Z=-4.736$, $p<.001$).} These results across $50$ participants are in line with \textbf{H2} and suggest that legible multi-robot teams encourage human participation.


\begin{figure}[t]
\centerline{\includegraphics[width=1.0\columnwidth]{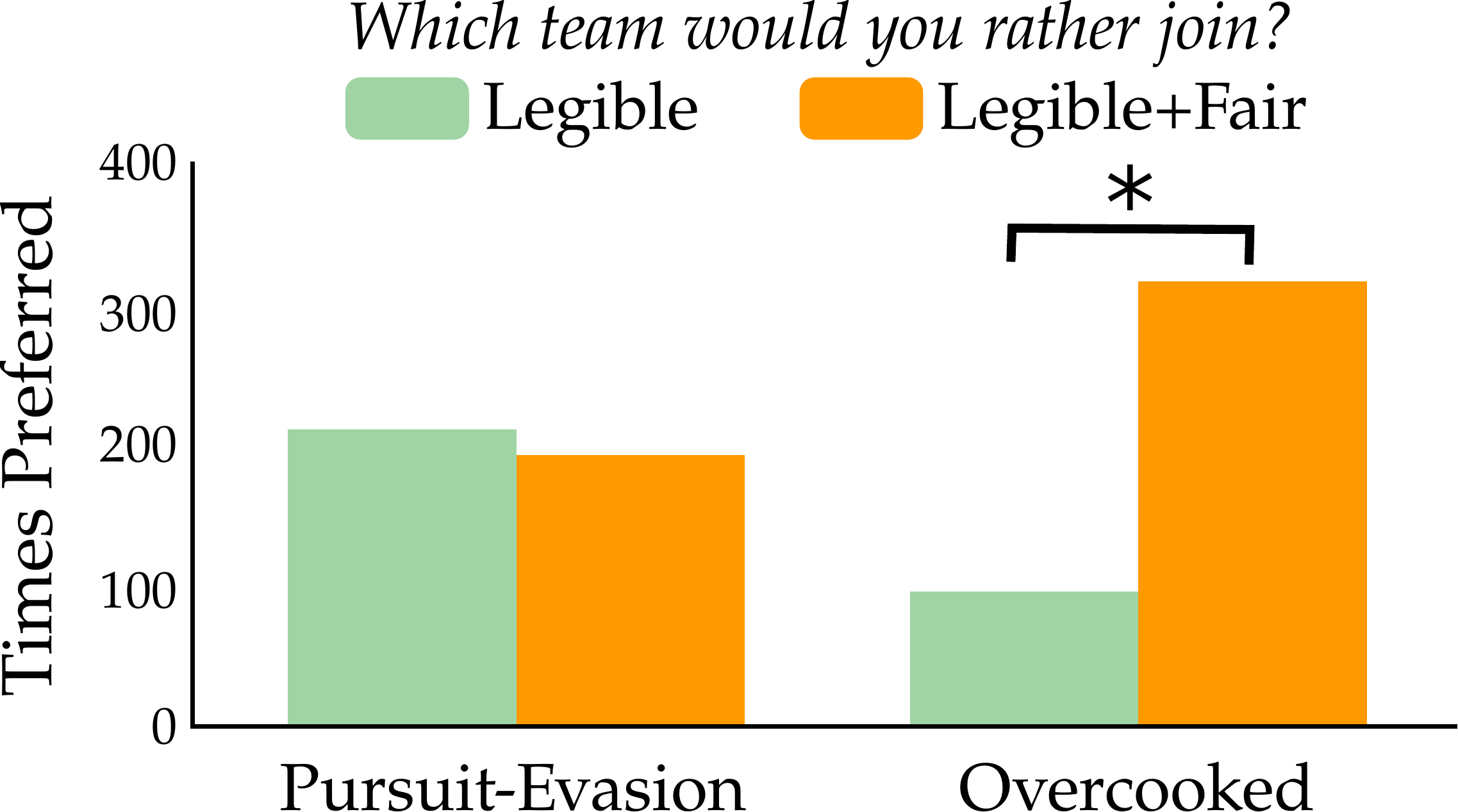}}
\caption{User preferences when watching \textbf{Legible} and \textbf{Legible+Fair} robot teams. Fifty participants watched eight pairs of teams and selected the teams they would prefer to join. We tested two definitions of fairness: in \textit{Pursuit-Evasion} the robots maintained equality of effort, and in \textit{Overcooked} the robots optimized for equality of allocation. {Participants preferred \textbf{Legible+Fair} teams in the \textit{Overcooked} environment ($p < .001$), but the differences were not statistically significant in \textit{Pursuit-Evasion} ($p=0.271$).}}


\label{fig:amt2}
\vspace{-1.5em}
\end{figure}

\begin{figure*}[t]
	\begin{center}
		\includegraphics[width=2.0\columnwidth]{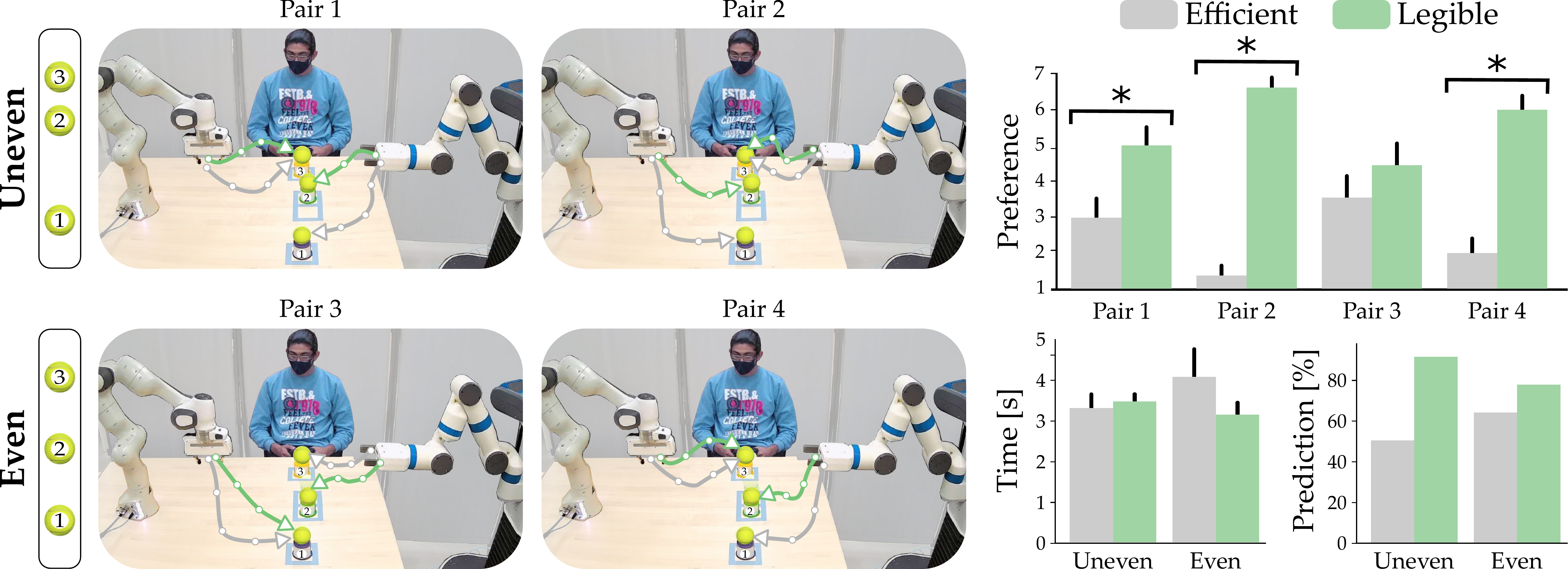}
		\caption{Comparing \textbf{Efficient} and \textbf{Legible} teams during our in-person user study. (Left) Participants collaborated with two $7$-DoF robot arms to clear tennis balls off the table. Users compared pairs of teams: in each pair one team optimized for completing the task efficiently, and the other optimized for making the human's role legible. {Note that the \textbf{Legible} allocations were not necessarily fair, and often the human had to reach across the table to complete their subtask.} (Right, Bottom) Participants more accurately predicted their role with \textbf{Legible} teams, but differences in prediction time were not statistically significant. (Right, Top) Participants preferred working with \textbf{Legible} teams. Here $*$ denotes statistical significance ($p<.05$).}
		\label{fig:user_study_leg}
	\end{center}
	\vspace{-1.5em}
\end{figure*}

\subsection{Fairness} \label{sec:watch_fairness}

The results of the first half of our \textit{watching} study indicate that legible robot teams convey allocations to the human observer, and that humans prefer to join these legible teams. But is the transparency of subtask allocations the only parameter that encourages human participation? Here we test the effects of \textit{fairness} when humans are watching robot teams.

\p{Independent Variables} Remember that our fair teams optimize for legibility in addition to fairness --- accordingly, we refer to this new condition as \textbf{Legible+Fair}. \textbf{Legible+Fair} teams used \eq{F3} to identify allocations that were fair for all agents. As a baseline, we compared these teams to the purely \textbf{Legible} allocations from the previous part. To isolate fairness, we selected \textbf{Legible} teams that \textit{were not} fair.

We also varied the definition of fairness used in \eq{F3}. Within \textit{Pursuit-Evasion} we defined fairness as equality of effort from \eq{F2}, and within \textit{Overcooked} we defined fairness as equality of allocation from \eq{F1}. In practice, \textbf{Legible+Fair} teams maintained an equal travel distance for each agent in \textit{Pursuit-Evasion}, and gave all agents an equal number of subtasks in \textit{Overcooked}.

\p{Procedure} Participants watched sixteen pairs of robot teams with different subtask allocations (eight pairs in \textit{Pursuit-Evasion} and eight pairs in \textit{Overcooked}). Each pair contained a \textbf{Legible+Fair} team and a \textbf{Legible} team. We randomized the order of the robot teams, and did not tell participants what objective each team was optimizing for. After watching each pair of teams participants selected the one they would prefer to collaborate with. Our hypothesis was that users would prefer to join teams that were both legible and fair:

\p{H3} \textit{Humans will prefer to join legible and fair robot teams where all members contribute equally to the task.}

\p{Results} The results of our second user survey are displayed in \fig{amt2}. As a reminder, here we are tallying the total number of pairs where participants selected \textbf{Legible} teams, and the total number of pairs where participants selected \textbf{Legible+Fair} teams. Interestingly, the results were not consistent across environments. Within \textit{Pursuit-Evasion} $47\%$ of the participants favored \textbf{Legible+Fair} teams, while in \textit{Overcooked} $77\%$ of the users preferred \textbf{Legible+Fair}. {Applying Wilcoxon signed-rank tests, the differences in participant preferences were not statistically significant in \textit{Pursuit-Evasion} ($Z=-1.100$, $p=.271$). By contrast, participants \textit{did} prefer to join \textbf{Legible+Fair} teams in \textit{Overcooked} ($Z=-10.7$, $p<.001$).}


To explain these results, recall that in the \textit{Pursuit-Evasion} environment we defined fairness as equality of effort. This means that \textbf{Legible} teams could cause members to reach for goals that were farther away. However, this distance did not seem to affect the human's perception: participants were just as willing to join teams where members had to travel unequal distances as teams that maintained equality of effort. By contrast, we focused on equality of allocation in \textit{Overcooked}. Here participants preferred \textbf{Legible+Fair} robot teams where all agents have an equal role --- i.e., people avoided teams where one or two chefs had to complete all the subtasks.

Our results partially support \textbf{H3}. People favored teams that optimized equality of allocation, but did not show a preference for teams that maintained equality of effort. This may have been because participants were only \textit{watching} teams and not actively \textit{playing} with those teams.


%% file: user-study.tex
\section{Encouraging Participation when Playing} \label{sec: user_study}

In Section~\ref{sec:mturk} we compared \textbf{Efficient}, \textbf{Legible}, and \textbf{Legible+Fair} allocations when humans were watching robot teams. Here we compare those same three approaches, but now with users that are \textit{actively participating} with the robot team. We repeat these studies because \textit{watching} is different from \textit{playing}: when humans watch teams they can consider the perspective of any agent; but when humans join in and collaborate with robot teams they must focus on their own allocation, and respond in real-time to the behavior of their robot teammates.

\p{Experimental Setup} Participants worked with two 7-DoF robot arms (Fetch, Fetch Robotics and Panda, Franka Emika). The two robots were centralized and shared a common controller. We placed three tennis balls within the workspace of the robot team: participants had to join in and help the robot team clear these tennis balls off the table (see \fig{user_study_leg}).

\p{Participants} We recruited 11 participants (3 female, 1 non-binary, ages 26 ± 3.3 years) from the Virginia Tech community. All participants provided informed written consent consistent with university guidelines (IRB $\#20$-$755$). {We used a within-subjects design: every participant interacted with \textbf{Efficient}, \textbf{Legible}, and \textbf{Legible+Fair} robots, and performed both parts of the user study described in Sections~\ref{sec:play_legibility} and \ref{sec:play_fairness}.}


\begin{figure*}[t]
	\begin{center}
		\includegraphics[width=2.0\columnwidth]{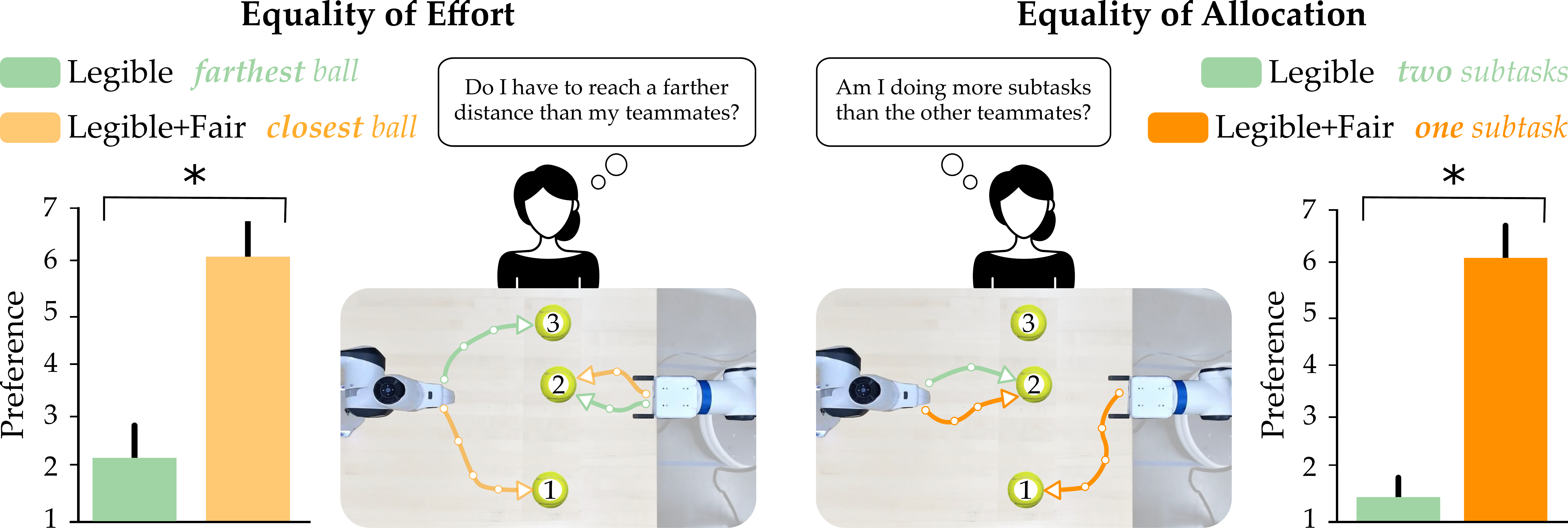}
		\caption{Comparing \textbf{Legible} and \textbf{Legible+Fair} teams during our in-person user study. We tested two definitions of fairness. (Left) Under equality of effort the \textbf{Legible+Fair} robot chose allocations where each agent travelled the same distance. (Right) Under equality of allocation the \textbf{Legible+Fair} robot chose allocations where the human was allocated one subtask. For both definitions users preferred \textbf{Legible+Fair} teams. Here $*$ denotes $p<.001$.}
		\label{fig:user_study_fair}
	\end{center}
	\vspace{-1.5em}
\end{figure*}

\subsection{Legibility}\label{sec:play_legibility}

In the first half of this user study we tested whether \textit{legibility} encourages humans to keep playing with robot teams. We compared efficient allocations (that minimize interaction time) to legible allocations (that communicate the human's role). We measured how accurately each participant was able to identify their subtask, as well as the participant's preferred team.

\p{Independent Variables} The robots leveraged \textbf{Efficient} and \textbf{Legible} allocations. For \textbf{Legible} the robots optimized \eq{L2} to select the allocation $\theta_{Leg}$ that best communicated the human's role. This is different from legibility when watching: instead of trying to make every agent's subtask clear, now the robots are \textit{only} trying to convey the participant's subtask (i.e., which tennis ball the human should remove from the table). To better isolate \textbf{Efficient} and \textbf{Legible}, we selected \textbf{Efficient} allocations that \textit{were not} legible.

\p{Procedure} Participants completed the cleaning task with four \textit{pairs} of robot teams (see \fig{user_study_leg}). Each pair contained one \textbf{Efficient} team and one \textbf{Legible} team. We never told the participants what type of team they were interacting with. To increase the number of data points, and to ensure that the position of the tennis balls did not affect our results, we placed the tennis balls in two configurations: \textit{Uneven} and \textit{Even}. In \textit{Uneven} balls $2$ and $3$ were clustered on one side of the table, while in \textit{Even} all of the tennis balls were equally spaced. During interaction the user needed to remove a ball from the table, but the user did not know \textit{a priori} which ball they should remove --- participants had to infer their assigned subtask from the actions of other agents. Participants sat next to the table and observed both robots' trajectories $\xi$. Once the participant was confident they knew which tennis ball to pick up, they pressed a button to temporarily pause the robots. After the user reached in and grabbed their tennis ball, the robots completed the rest of the task autonomously.

\p{Dependent Measures} When the human was interacting with a robot team we recorded the amount of \textit{time} that user waited before intervening, and whether or not the user \textit{predicted} their subtask correctly. After the participant finished interacting with a pair of teams (e.g., Team A and Team B) participants indicated which team they would prefer to keep collaborating with on a scale of $1$-$7$. Here a score of $1$ denotes that the human had a strong preference for Team A, while a score of $7$ denotes that the human had a strong preference for Team B.

\p{Hypotheses} We had two hypotheses:

\p{H4} \textit{Humans that play with legible robot teams will more quickly and accurately recognize their role within the team.}

\p{H5} \textit{When given the choice, humans will prefer to keep playing with legible robot teams.}

\p{Results} Our results are summarized in \fig{user_study_leg}. First, we conducted a repeated measures ANOVA to confirm that the arrangement of the tennis balls \textit{did not} have an effect on the human's preferences ($F(1, 21) = .000, p = 1.00$). Next, we counted the number of times the participants correctly predicted their subtasks with \textbf{Efficient} and \textbf{Legible} teams: similar to our online study in \fig{amt_pred}, participants made the correct prediction more frequently when working with \textbf{Legible} teams. {Interestingly, the type of robot team did not have an effect on the time it took for humans to make these predictions ($F(1, 21) = 1.843$, $p= .189$)}. The combination of these results partially supports \textbf{H4} --- legible task allocations led to more accurate human predictions, but not faster predictions.

Finally, we analyzed whether participants preferred to keep playing with \textbf{Efficient} or \textbf{Legible} teams. We conducted a one-way repeated measures ANOVA with a Sphericity Assumed correction and determined that the type of allocation (\textbf{Efficient} or \textbf{Legible}) had a significant main effect on the users’ preference: $F(1, 10) = 77.13, p < .001$. In all four pairs we found higher average scores for \textbf{Legible}, and in three out of the four pairs our post hoc t-tests showed that this difference was statistically significant ($p<.05$). Although users scored \textbf{Legible} higher in Pair $3$, here the difference between \textbf{Efficient} and \textbf{Legible} was not statistically significant ($t(20) = -1.056, p=.303$). These results are consistent with \textbf{H5}, and suggest that people prefer to continue collaborating with teams that make their roles clear.

\subsection{Fairness}\label{sec:play_fairness}

Our results so far suggest that humans prefer to interact with robot teams that optimize for legible (but potentially inefficient) allocations. Next, we test how adding \textit{fairness} into these allocations changes the human's preferences.

\p{Independent Variables} {We compared \textbf{Legible+Fair} teams that optimize \eq{F4} to \textbf{Legible} teams which follow the same approach as in Section~\ref{sec:play_legibility}}. To better separate these conditions we purposely selected \textbf{Legible} teams that were not fair. Recall that \eq{F4} optimizes for the fairness $f_{\mathcal{H}}$ \textit{from the human's perspective}. This incentives the robots to select allocations where the human has an equal share of the work, and does not penalize the robots for dividing the remaining effort unevenly among themselves (e.g., a single robot may be asked to pick up multiple tennis balls under \textbf{Legible+Fair}).

Similar to our \textit{watching} user study in Section~\ref{sec:mturk}, we studied two different definitions of fairness in teams: \textit{equality of allocation}, \eq{F1}, and \textit{equality of effort}, \eq{F2}. Examples of the trajectories generated by these robot teams are shown in \fig{user_study_fair}. Under equality of effort the robots assigned the human to closest ball, and under equality of allocation the robots always performed two of the three subtasks.

\p{Procedure} Participants collaborated with two pairs of teams to clear the table. In one pair the \textbf{Legible} team asked users to pick up the farthest ball while the \textbf{Legible+Fair} team asked participants to get the closest ball. In the other pair the \textbf{Legible} team asked participants to pick up two balls while the the \textbf{Legible+Fair} team asked users to get one ball. After interacting with a pair of teams users indicated their preference on a $1$-$7$ scale (just as in Section~\ref{sec:play_legibility}). We hypothesized that:


\p{H6} \textit{Participants will prefer to join robot teams that legibly and fairly distribute the allocations or effort.}


\p{Results} Our results in \fig{user_study_fair} indicate that humans have a preference for fairness when they are playing with robot teams. Across both definitions of fairness, participants rated \textbf{Legible+Fair} teams significantly higher than \textbf{Legible} teams (one-way repeated measures ANOVA: $F(1,10) = 40.06, p<.001$). We contrast these results to \fig{amt2}, where the watching humans marginally preferred \textbf{Legible} teams in Pursuit-Evasion. Comparing these results, we suggest that fairness may be less of a factor when users are watching the team, but more decisive when humans are actually playing with the team.



%% file: conclusions.tex
\section{Conclusions} 

We developed an optimization framework that enables teams of robots to encourage human participation. Under our approach centralized robot teams treat humans as humans, and actively search for legible and fair ways to allocate subtasks among agents. As compared to a baseline that purely optimizes for efficiency, robots that leverage legible and fair allocations better encourage \textit{watching} humans to join the team and \textit{playing} humans to keep collaborating with the team.